\documentclass[journal]{IEEEtran}

\usepackage{cite}
\usepackage{amsmath,amssymb,amsfonts}
\usepackage{booktabs}
\usepackage{tabularx}
\usepackage{array}
\usepackage{graphicx}
\usepackage{algorithm,algorithmic}
\usepackage{textcomp}

\graphicspath{{figures/}{./}}

\newcommand{\etal}{\textit{et al.}}
\DeclareMathOperator{\ECE}{ECE}

\usepackage{hyperref}
\hypersetup{
  hidelinks,
  pdfencoding=unicode,
  pdftitle={Auditing Medical Vision\textendash Language Models on Chest Radiographs: Estimating Reference Agreement Across Institutions},
  pdfauthor={Pengyang Yu, Yiou Wang, Zhongping Dong, Sahraoui Dhelim, Chun-Mei Feng, and M. Tahar Kechadi},
  pdfsubject={IEEE Transactions on Medical Imaging Special Issue manuscript},
  pdfkeywords={Chest radiography; external validation; medical vision-language models; multi-site study; reference-agreement estimation; uncertainty evaluation}
}

\begin{document}

\title{Auditing Medical Vision--Language Models on Chest Radiographs:
Estimating Reference Agreement Across Institutions}

\author{Pengyang~Yu, Yiou~Wang, Zhongping~Dong,
        Sahraoui~Dhelim, Chun-Mei~Feng, and~M.~Tahar~Kechadi%
\thanks{This work was supported by University College Dublin.}%
\thanks{Pengyang~Yu, Zhongping~Dong, Chun-Mei~Feng, and M.~Tahar~Kechadi are with the School of Computer Science, University College Dublin, Dublin, Ireland (e-mail: pengyang.yu@ucdconnect.ie; zhongping.dong@ucdconnect.ie; chunmei.feng@ucd.ie; tahar.kechadi@ucd.ie).}%
\thanks{Yiou~Wang is with the Department of Medical Imaging, The Third Affiliated Hospital of Southern Medical University, Guangzhou, China (e-mail: wyo22321320@smu.edu.cn).}%
\thanks{Sahraoui~Dhelim is with Dublin City University, Dublin, Ireland (e-mail: sahraoui.dhelim@dcu.ie).}%
\thanks{Corresponding authors: Pengyang~Yu and Chun-Mei~Feng.}%
}

\maketitle

\begin{abstract}
Vision--language models return structured chest-radiograph findings through interfaces exposing no confidence score, so a receiving institution cannot read off how far to trust an individual judgment. Whether agreement with an institution's reference standard transfers across sites, findings, prediction directions and question formats is largely unmeasured. We evaluated three generative vision--language models on three institutional chest-radiograph corpora and six findings under two elicitation protocols, comprising more than 345{,}000 finding-level predictions, and estimated finding-by-direction reference agreement at a receiving institution from a small budget of local labels. Estimation strategies were then stress-tested under repeated strict institution-held-out evaluation. Under evaluation excluding the receiving institution from development entirely, adaptive selection among the seven estimators that design admits did not improve on simple fixed alternatives: it achieved a mean Brier score of 0.1083, against 0.0853 for always using a Beta--Binomial empirical-Bayes estimator and 0.0855 for a target-only logistic model. Those two differ by 0.0003, less than this family's own sensitivity to a change of solver version, and each leads in about half the settings, so no default can be recommended. Their advantage over estimators pooling across institutions was concentrated at one site and not confirmatory once clustered by institution, and a plug-in empirical-Bayes posterior-predictive count interval at a nominal 95\% level covered 87.0\%, less at the hardest institution. Reference agreement therefore has to be re-evaluated per site and per interface; these results concern agreement with institutional labels, not clinical correctness.
\end{abstract}


\section{Introduction}
\label{sec:introduction}

\IEEEPARstart{M}{edical} vision-language models (VLMs)~\cite{multimodal_med_foundation} now produce structured finding-level judgments for chest radiographs and report competitive performance on selected benchmarks~\cite{medgemma,chexagent}, but the interfaces through which they are consumed commonly return a categorical verdict and nothing else---no score, no internals---so an institution adopting one has no direct way to know how far a particular judgment can be relied upon.

Current evidence does not close that gap: hallucination benchmarks and domain-shift studies alike report scores in aggregate. Neither answers what an adopting hospital needs: when the model is moved, do the judgments it gets wrong change predictably, and can the receiving institution measure that itself? What is measurable is agreement with the institution's own reference standard---the estimand throughout, not clinical correctness.

\begin{figure*}[t]
\centering
\includegraphics[width=\textwidth,trim=0 122.3 3.8 0,clip]{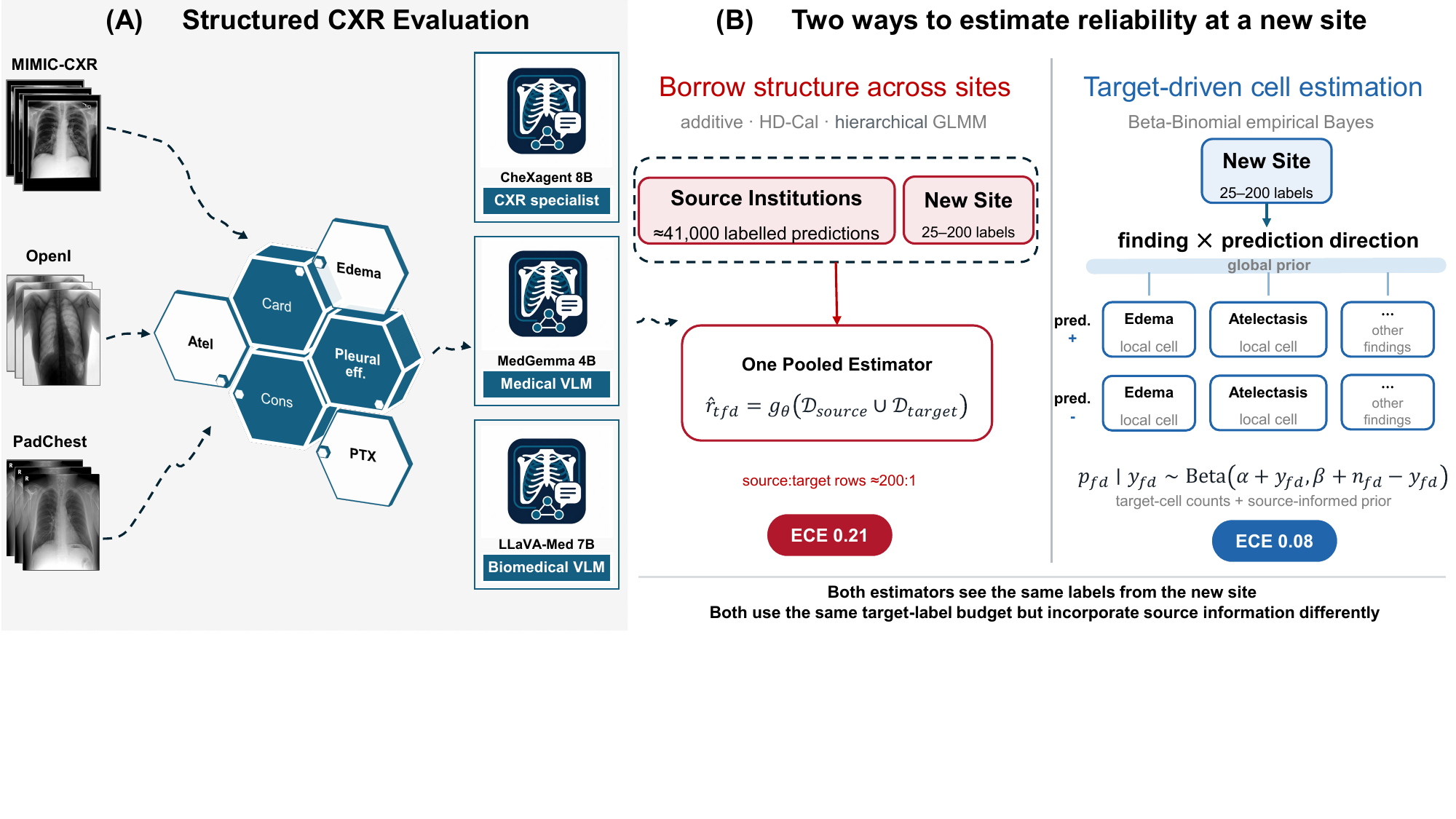}
\caption{Study design and the two families of reliability estimator. (A) Three generative VLMs on three institutional datasets (3{,}066, 3{,}851, and 4{,}998 evaluable studies) and six findings, under two elicitation protocols, yielding 165{,}432 narrative and 180{,}210 binary finding-level predictions. (B) Pooling estimators share parameters with the source institutions, which outnumber the target sample; Beta--Binomial instead updates each target cell from its own counts under a prior estimated from the same pool. Both receive an identical label budget, so the families differ in how source information enters, not how much they see. The two calibration errors are one worked deployment---MedGemma 4B at MIMIC-CXR with 200 target labels---not a mean over the 24 settings, for which see Table~\ref{tab:hdcal_vs_baselines}.}
\label{fig:overview}
\end{figure*}

We study three generative VLMs---MedGemma 4B~\cite{medgemma}, CheXagent 8B~\cite{chexagent} and LLaVA-Med 7B~\cite{llava_med}---on MIMIC-CXR~\cite{mimiccxr}, OpenI~\cite{openi} and PadChest~\cite{padchest} across six findings and two elicitation protocols. Spanning two countries, two languages and three label-generation regimes, they form a heterogeneous retrospective stress test under three-fold external validation (Fig.~\ref{fig:overview}).

We make three contributions. The first characterizes how reference agreement varies with model, institution, finding, prediction direction and elicitation protocol: after Benjamini--Hochberg adjustment over all 36 contrasts per metric, 23 false-positive-rate and 24 false-negative-rate contrasts remain significant, the institution effect persists with the finding held fixed, and the two primary VLMs, on identical images and labels, agree on about a third of the significant contrasts (Jaccard 0.35).

The second is a strict external validation of the estimation strategies under a design excluding the receiving institution from development altogether, not merely the deployment being served: adaptive selection among the seven estimators that design admits does not improve on the simple fixed alternatives, whose own ordering is unstable across settings. The third is an uncertainty audit treating interval coverage as an endpoint rather than a construction detail: an interval aligned with what a held-out fold can test does not attain its nominal level, the shortfall concentrating at the hardest institution and in predictions asserting a finding.

\section{Related Work}
\label{sec:related_work}

\subsection{Medical VLMs, Hallucination, and Domain Shift}

General medical VLMs include LLaVA-Med~\cite{llava_med} and MedGemma~\cite{medgemma}; CheXagent~\cite{chexagent} is a chest radiograph specialist with a structured binary interface, all within the wider programme of generalist medical foundation models~\cite{multimodal_med_foundation,radfm}. UniChest~\cite{unichest} pre-trains across several sources against their heterogeneity, but targets training generalization rather than deployment-time behaviour. Hallucination has been approached through hidden-representation detection~\cite{rextrust}, sampling-based flagging~\cite{radflag} and dedicated benchmarks~\cite{medvh,crest,cares}, none of which compares matched model-by-finding error patterns across institutional deployment environments under a common protocol. Cross-institutional degradation has been documented since Zech and colleagues~\cite{domainshift_medical,finlayson2021}, but that literature reports aggregate metric drops; we ask whether the \emph{structure} of the error changes along the institution-by-finding axis.

\subsection{Uncertainty Methods and Their Assumptions}

Post hoc calibration~\cite{guo2017calibration} requires continuous scores and degrades under distribution shift~\cite{ovadia2019}, and selective~\cite{geifman_selective,selectivenet} and conformal prediction~\cite{conformal,conformal_risk} likewise presume a score to threshold or rank. The interfaces studied here return a categorical verdict whose content depends also on the prompt, the decoding and the parser, removing that assumption. Concurrent work on a medical VLM shifted from MIMIC-CXR to PadChest finds simple single-model uncertainty outperforming ensembling~\cite{sadanadan2026}, but operates on token logits. Partial pooling is likewise classical: empirical Bayes shrinkage of binomial cell rates and hierarchical generalized linear models are standard instruments, used here unchanged. What is new is the evaluation---a strict institution-held-out stress test, a direct comparison of adaptive selection with fixed policies, and interval coverage measured as an empirical endpoint.

\section{Materials and Methods}
\label{sec:methods}

\subsection{Datasets and Deployment Environments}
\label{sec:datasets}

We analyzed three public chest radiograph datasets representing distinct deployment environments, with their cohort attrition in Table~\ref{tab:datasets}.

\begin{table}[t]
\centering
\caption{Attrition from sampled studies to scored judgments, which are evaluable study--finding--model records after reference-label eligibility and parsing. A finding is scored only when it carries a definite present or absent reference label, and narrative elicitation additionally requires four of six findings to parse, so judgment counts differ by model. Prevalence is the range across the six findings.}
\label{tab:datasets}
\footnotesize
\setlength{\tabcolsep}{3.5pt}
\renewcommand{\arraystretch}{1.15}
\begin{tabularx}{\columnwidth}{@{}l X r@{\,$\to$\,}r r c@{}}
\toprule
 & Label & \multicolumn{2}{c}{Studies} & Judgments & Prev- \\
Domain & source & \multicolumn{2}{c}{sampled\,/\,evaluable} & MedG./CheX. & alence \\
\midrule
MIMIC-CXR & CheXpert    & 5{,}000 & 3{,}066 & 3{,}912/6{,}966   & 0.19--0.97 \\
OpenI     & English NLP & 3{,}851 & 3{,}851 & 19{,}541/23{,}093 & 0.004--0.097 \\
PadChest  & Spanish NLP & 5{,}000 & 4{,}998 & 21{,}862/29{,}988 & 0.001--0.093 \\
\bottomrule
\end{tabularx}
\end{table}

\textbf{MIMIC-CXR}~\cite{mimiccxr}, from an academic intensive-care population and labelled by the CheXpert automated labeler~\cite{chexpert_labeler}, contributes the high-prevalence, high-acuity environment: $5{,}000$ studies (seed 42) from $3{,}918$ distinct patients, without patient-level deduplication. Since the protocol assigns whole datasets to a role, repeated patients cannot cross institutions, and the residual clustering is quantified in Section~\ref{sec:dedup_sensitivity}. \textbf{OpenI}~\cite{openi}, from a United States national research hospital, contributes a routine outpatient population at much lower prevalence. \textbf{PadChest}~\cite{padchest}, from a Spanish regional hospital, adds cross-national and label-pipeline variation: its native 174-label taxonomy was mapped to our six findings by an explicit dictionary with exclusion rules (pericardial effusion was excluded from pleural effusion), and we sampled $5{,}000$ patient-deduplicated studies with seed 42, excluding pediatric cases. The model receives only the image and an English prompt. The six findings are atelectasis, cardiomegaly, consolidation, edema, pleural effusion and pneumothorax.

\subsection{Vision-Language Models}
\label{sec:models}

\textbf{MedGemma 4B}~\cite{medgemma} (\texttt{google/medgemma-4b-it}) is a general medical VLM of about 4 billion parameters. Under narrative elicitation it was asked for an explicit present or absent statement per finding, and judgments were extracted by a deterministic parser with a global-normal fallback for reports asserting overall normality; a study was evaluable if at least four of six findings parsed, at a parse success rate of 99.93\%. \textbf{LLaVA-Med 7B}~\cite{llava_med} (\texttt{microsoft/llava-med-v1.5-mistral-7b}), on a different language backbone, used the identical prompt and parser. \textbf{CheXagent 8B}~\cite{chexagent} (\texttt{StanfordAIMI/CheXagent-8b}), a chest radiograph specialist of about 8 billion parameters, exposes a structured binary interface returning a Yes or No verdict per finding, so no parsing is required. The three checkpoints were frozen at the Hugging Face revisions \texttt{290cda5e} (MedGemma), \texttt{4934e914} (CheXagent) and \texttt{91bb16c1} (LLaVA-Med).

Under the \emph{binary} protocol each finding was requested separately and a single token generated, recording the logits of the affirmative and negative responses. All three VLMs were run under both protocols.

Two filters define the evaluable set. Across the two primary models, 13 of $27{,}702$ study--model responses ($0.047\%$) were unparseable and excluded. Reference-label eligibility was then applied separately to each study--finding pair, so a retained study did not necessarily contribute all six findings. That second filter binds very unevenly (Table~\ref{tab:datasets}), MIMIC-CXR's labeler recording an uncertain or unmentioned label far more often: at MIMIC-CXR it cuts MedGemma from $18{,}547$ parsed finding-level rows to $3{,}912$ and CheXagent from $29{,}959$ to $6{,}966$, leaving between 221 and $1{,}274$ scored studies per finding there, while at OpenI and PadChest every parsed row already carried a definite label. How a corpus assigns uncertain and unmentioned mentions to that split is recorded for MIMIC-CXR, through the CheXpert labeler's four-way output, and for PadChest, through the explicit dictionary above; for OpenI it is not recorded in the retained artifacts. What is estimated throughout is therefore reliability on the labelled subset. The two primary VLMs contribute $105{,}362$ narrative predictions and LLaVA-Med a further $60{,}070$. The estimator comparison uses the narrative predictions of the two primary VLMs, where all ten estimators are defined.

\subsection{HD-Cal Formulation}
\label{sec:hdcal}

Let $m$ denote a VLM, $d$ a deployment domain, $k$ a clinical finding, and $r\in\{\text{present},\text{absent}\}$ the direction of a structured judgment $\hat y$. With reference label $y$, the estimation target is the binary correctness $c=\mathbb{1}[\hat y = y]$---agreement with the institution's reference standard, not disease presence---HD-Cal models that target as a logistic function of one-hot indicators for $[r,k,d,d{:}k]$,
\begin{equation}
\begin{split}
P(c=1\mid m,d,k,r)=\sigma\big(&\beta_0^{(m)}+\beta_r^{(m)}+\beta_d^{(m)}\\
&+\beta_k^{(m)}+\beta_{d,k}^{(m)}\big),
\end{split}
\label{eq:hdcal}
\end{equation}
fitted per VLM by $\ell_2$-regularized maximum likelihood ($C=1$, L-BFGS). The interaction $\beta_{d,k}^{(m)}$ encodes the heterogeneity of Section~\ref{sec:discovery}: if the structure lies on the institution-by-finding axis, that term captures variance no marginal specification can.

\subsection{Baseline Ablation}
\label{sec:baselines}

We compared ten estimators in three families, by how much information they carry across institutions. The \emph{pooled-rate constant} (implementation key \texttt{naive}) assigns every test judgment the mean agreement in that fold's fitting pool---the available source rows plus the target-label shot. It uses local labels at non-zero budgets but no finding-, direction- or institution-specific structure; at zero shots it reduces to the pooled source mean. The second pools source and target data under shared parameters, so the sources necessarily influence the target estimate: the \emph{additive} model ($r{+}d{+}k$); \emph{HD-Cal} ($r{+}d{+}k{+}d{:}k$), adding the institution-by-finding interaction; a \emph{weighted} variant up-weighting target rows by the source-to-target size ratio; a \emph{hierarchical GLMM} with half-normal priors on the finding, institution and interaction variances, so that the amount of pooling is learned rather than fixed; and an \emph{adaptive-shrinkage} estimator (HAS) designed for this study, taking the source-fitted per-cell reliability as a fixed offset and estimating a site-level shift plus finding-specific deviations from the target labels alone, each shrunk by a strength derived from that finding's variability between source institutions. Two ablations fix that strength or drop the offset. The GLMM uses an aggregated Binomial likelihood with a logit link, $N(0,2.5)$ priors on the intercept and the direction coefficient, and crossed finding, institution and institution-by-finding random effects with $\mathrm{HalfNormal}(1)$ hyperpriors on their standard deviations; it was fitted with an AutoNormal guide, Adam($0.02$) and $3{,}000$ SVI steps, and plug-in predictions use the guide's constrained latent-site medians, an unseen interaction contributing zero. The weighted variant gives source rows weight one and target-shot rows weight $n_{\mathrm{source}}/n_{\mathrm{shot}}$ with no user-side normalization. HAS takes a direction-stratified source-cell offset and shrinks finding-specific deviations by $\tilde\tau^2_a=\mathrm{clip}(0.5\hat\tau^2_a+0.5\overline{\hat\tau^2},0.05,25)$, where $\hat\tau^2_a$ is half the squared contrast between the two source institutions' smoothed logits for finding $a$, pooled over prediction directions; the fixed-$\tau^2$ ablation replaces $\tilde\tau^2_a$ by its fold mean, and the no-offset ablation drops the offset while keeping the source-derived $\tilde\tau^2$. Because $\hat\tau^2$ needs two source institutions, all three are undefined when only one is available (Section~\ref{sec:protocol}).

The third estimates each cell from the target labels. A \emph{target-only logistic} model fits the full specification to the $n_{\mathrm{shot}}$ labels with no source data, isolating the value of shrinkage. \emph{Constrained method-of-moments Beta--Binomial empirical Bayes} treats each (institution, finding, direction) cell as a binomial sample: with $y_j$ correct of $n_j$ judgments in cell $j$, its posterior mean is
\begin{equation}
\hat P_j=\frac{y_j+\alpha}{n_j+\alpha+\beta},
\qquad \alpha=\hat\mu\hat\kappa,\quad \beta=(1-\hat\mu)\hat\kappa,
\label{eq:bb}
\end{equation}
\begin{equation}
\hat\kappa=\max\!\left\{\frac{\hat\mu(1-\hat\mu)}{\max(\hat v,10^{-6})}-1,\;1\right\}.
\label{eq:kappa}
\end{equation}
Here $\hat\mu$ and $\hat v$ are the mean and variance of the observed per-cell agreement rates, each weighted by that cell's judgment count, over every cell of the fitting pool---the two source institutions with the target's labelled sample, so source cells dominate by roughly two hundred to one. $\hat v$ is the raw weighted variance and does not subtract the binomial sampling component, so it over-states between-cell variance; $\hat\kappa\ge1$ and the guard on the denominator are stability conditions, not part of the moment estimator. The prior is therefore source-informed, only the counts $y_j$ and $n_j$ of a target cell come from the target labels, and an absent cell falls back to the pooled rate. The three families thus span the borrowing spectrum, and under an identical label budget the comparison across them is between borrowing structure and estimating each cell from local counts.

\subsection{Evaluation Protocol}
\label{sec:protocol}

Reporting standards prescribe what a validation study must disclose~\cite{claim2024,tripod_ai}, but not how a receiving institution should obtain a reliability estimate for a model it did not develop and cannot inspect. Algorithm~\ref{alg:protocol} states one candidate answer, and is the object this study evaluates rather than a procedure we ask an institution to adopt. It presumes only a categorical interface, requires neither internals nor retraining, and is explicit about the two steps most easily skipped: a gate asking whether any estimator improves on the constant predictor, and selection on a proper score.

\begin{algorithm}[t]
\caption{Deployment-time reliability estimation.}
\label{alg:protocol}
\begin{algorithmic}[1]
\STATE \textbf{Given} a VLM reachable only as a categorical interface, a receiving institution $d^\star$, findings $\mathcal{K}$, a budget of $n$ labels, and candidate estimators $\mathcal{E}$.
\STATE Draw $n$ finding-level judgments from prior studies at $d^\star$, stratified over $\mathcal{K}$; obtain a reference label for each, record the model's judgment $\hat y$ and direction $r$, and form $c=\mathbb{1}[\hat y=y]$.
\STATE \emph{Fit-for-purpose gate.} Score every $e\in\mathcal{E}$ and the constant predictor under leave-one-institution-out validation; if none beats the constant, report the constant and stop.
\STATE \emph{Selection.} Choose $e^\star$ by a proper scoring rule, breaking ties by a fixed candidate order declared in advance.
\STATE \emph{Reporting.} Give per-cell $\hat P(c{=}1)$ with intervals, the deployment-level margin over the constant predictor, and the solver version.
\STATE \emph{Revalidation.} Return to step 2 whenever the model, the institution, or the elicitation protocol changes.
\end{algorithmic}
\end{algorithm}

The study instantiates Algorithm~\ref{alg:protocol} under a leave-one-institution-out design, letting each institution in turn stand as the receiving site. For each target the two remaining institutions form the source set, to which $n_{\mathrm{shot}}\in\{0,25,50,100,200\}$ uniformly sampled target labels are added, the remainder serving as the test set; each setting was repeated ten times from a seed base of 42. The budget counts finding-level judgments, not studies, though at these sizes the drawn rows are almost all distinct studies---25, 50, 98--100 and 194--196 at the four budgets. Fit and test rows are disjoint, but a study contributing several scored findings can place some on each side; Section~\ref{sec:sensitivity} repeats the comparison under a study-disjoint split, and with no patient identifier in the prediction tables disjointness holds at study level only. Logistic-family estimators used L-BFGS, $C=1.0$, at most $1{,}000$ iterations, and an unweighted likelihood. At zero shots the structured estimators reduce to their marginal counterparts, so the 24 settings are the six deployments crossed with the four non-zero budgets.

Evaluating the selection step calls for a stricter hold-out than withholding the deployment being served: two models are deployed at each institution, so that would leave the receiving site represented by its other model. We exclude the receiving institution $I$ entirely. Development runs inside the remaining pair---each of the other two institutions in turn acts as a pseudo-target $J$, whose complete source data comes from the single remaining institution $K$ while $J$ contributes its own local label shot---so no row, prior or weight derived from $I$ enters a development fit or its scoring. Algorithm~\ref{alg:protocol} begins from a given VLM, so a decision is made for one model at a time: candidates are ranked by mean Brier over the \emph{two same-model} development folds---the two pseudo-targets, scored for that model only---at the same budget and repetition. No score from the other model enters the decision. The gate is applied to the same evidence, returning the pooled-rate constant unless the best candidate is \emph{strictly} lower, and exact ties break toward the earlier entry of a fixed candidate order. A decision is therefore indexed by institution, model, budget and repetition---$3\times2\times4\times10=240$ decisions---each applied to its own model-institution deployment, giving 240 target-deployment evaluations, which are not independent. The selected estimator is then fitted at $I$ exactly as any fixed policy is, on the two non-$I$ sources plus $I$'s own label shot, and scored on $I$'s held-out complement.

Two candidate sets follow, and are not interchangeable. \emph{Set A}, the fixed-policy benchmark of Table~\ref{tab:hdcal_vs_baselines}, is all ten estimators, each fitted on two complete source institutions plus the target shot. \emph{Set B}, the selector's action set, is the seven defined on a single-source development fold; the three adaptive-shrinkage variants are not, building their shrinkage strength from the contrast between two source institutions. That exclusion is structural, not performance-based, and was fixed before any outcome under this design was computed---but Set B is \emph{post-hoc}: not pre-registered, and not a ten-candidate selector. Logistic-family estimators are reference-coded, with a fitted \emph{unpenalized} intercept and $\ell_2$-penalized feature coefficients; a single-class fitting pool falls back to the pooled-rate constant. The Set-B \emph{per-cell oracle} takes the best of the seven in each cell after the fact: an unattainable lower bound rather than a policy, so differences against it are relative excess Brier and not regret, the two not ranging over a matched action space. Five of the ten are scorable under the binary protocol here, so the contrast between interfaces runs on that matched set of fixed policies, post-hoc and exploratory.

Step 5 asks for an interval, and the audit of it is specific to one estimator---and to what a held-out fold can test. Equation~(\ref{eq:bb}) yields a credible interval for a cell's latent agreement probability $p_j$, which the folds never reveal, so containment of a noisy empirical proportion is not a coverage statement about it in either direction. We therefore audit the held-out count itself: with $S_{\mathrm{train}}$ of $N_{\mathrm{train}}$ target-training judgments correct in cell $j$ and the same source-informed prior, $Y_{\mathrm{test}}\mid\mathrm{data}\sim\mathrm{BetaBin}(N_{\mathrm{test}},\,\alpha+S_{\mathrm{train}},\,\beta+N_{\mathrm{train}}-S_{\mathrm{train}})$, from which we take the exact equal-tailed $95\%$ interval $[L,U]$ on the integer quantiles. It covers when $L\le Y_{\mathrm{test}}\le U$, its width is $(U-L)/N_{\mathrm{test}}$ on the probability scale, and a cell is eligible when it appears in the fitting pool with $N_{\mathrm{test}}\ge20$. Nothing else moves: $\hat\kappa$, its floor, the guarded variance, the counts and the test definition are those of (\ref{eq:bb})--(\ref{eq:kappa}). These evaluations overlap heavily---the same cell recurs across budgets and repetitions---so coverage is descriptive repeated-fold empirical coverage, without an interval of its own. No comparable interval is defined here for the other candidates, so the result does not validate the intervals of a selected estimator in general.

\subsection{Statistical Procedure}
\label{sec:stats}

Differences in false positive and false negative rate were tested with Fisher exact tests over the $6\times3\times2=36$ institution-pair contrasts per metric. Both reference-negative denominators were non-zero in every false-positive-rate contrast; seven contrasts had zero false-positive events at both institutions, yielding $p=1$, and were retained. Benjamini--Hochberg adjustment was applied separately to all 36 false-positive-rate and all 36 false-negative-rate $p$ values at $\alpha=0.05$~\cite{bh1995}.

The correction family is defined over the two primary VLMs. LLaVA-Med is reported separately because, under narrative elicitation, its parsed outputs did not vary with the image (Section~\ref{sec:elicitation}).

Cochran--Mantel--Haenszel tests stratified by finding were run for the six model-by-institution-pair combinations; all six findings gave admissible strata for CheXagent, while for MedGemma strata collapsing to empty tables were dropped, leaving four and three---a conservative reduction, those strata carrying no signal. Intervals for cell-level rate differences are bootstrap confidence intervals resampling studies as clusters, $10{,}000$ iterations. Cross-model overlap is the Jaccard index between each model's set of significant contrasts.

Estimator comparisons over the 24 model-by-institution-by-budget settings are summarized descriptively after averaging each setting over its ten repetitions. The settings share institutions, models, studies and nested label budgets, so no setting-level inferential test is used, and the interval on the headline difference between estimator families is resampled over institutions rather than deployments (Section~\ref{sec:calibration_main}).

A stratified single-reader audit covered 150 studies (50 per institution) from four strata: suspected false positives, suspected false negatives, high-risk cells, and representative cases. One reviewer, a doctoral researcher in medical imaging, labelled each study for the six findings as present, absent, or uncertain, blinded to model outputs and risk category. It is a preliminary descriptive audit: agreement with the dataset labels is summarized by Cohen's $\kappa$ over the unique study-by-finding judgments, and the risk-tertile comparison is given as counts and a difference, with no inferential test, the units being model-expanded and cluster-aware procedures not agreeing on it.

This retrospective secondary analysis used previously collected, de-identified chest-radiograph datasets. Datasets requiring credentialed access were accessed only by authors who had completed the applicable authorization requirements and held the required dataset-specific access permissions and data-use agreements; the remaining datasets were publicly available. The single-reader audit was performed by co-author Yiou Wang as a member of the research team, rather than by an external research participant. The study involved no new patient recruitment, patient contact, clinical intervention, or attempted re-identification, and no new patient consent was collected for this secondary analysis.

\subsection{Evaluation Metrics}
\label{sec:metrics}

The primary calibration metric is the expected calibration error (ECE) with $B=10$ equal-width bins~\cite{naeini2015}: predictions are binned by their estimated reliability and the sample-weighted absolute gap between empirical accuracy and mean estimated reliability is summed,
\begin{equation}
\ECE = \sum_{b=1}^{B} \frac{|S_b|}{N} \left| \mathrm{acc}(S_b) - \mathrm{conf}(S_b) \right|,
\label{eq:ece}
\end{equation}
over the $B$ bins $S_b$ of a test set of size $N$, with $\mathrm{acc}(S_b)$ the empirical accuracy in bin $b$ and $\mathrm{conf}(S_b)$ the mean estimated reliability there. Binned ECE is biased and sensitive to the binning scheme~\cite{kumar2019}, so the Brier score $\mathrm{BS}=N^{-1}\sum_i(\hat P_i - c_i)^2$ is reported alongside it as a binning-free proper score, and settles selection when the two disagree.

\subsection{Implementation and Reproducibility}
\label{sec:reproducibility}

VLM inference ran on NVIDIA A100 GPUs in half precision (\texttt{bfloat16}). Logistic-family estimators used \texttt{scikit-learn} 1.2.2~\cite{sklearn} with L-BFGS; the hierarchical model \texttt{NumPyro} 0.16.1 on \texttt{JAX} 0.4.35; tests \texttt{scipy} 1.15.3. Seeds were fixed at a base of 42, the ten repeats using 42 through 51.

We report solver versions because they matter here. Re-running the logistic family under \texttt{scikit-learn} 1.9.0 reproduced the constant and Beta--Binomial estimators to machine precision but diverged from the reported values by up to 0.23 calibration error, worst at the smallest budgets on MIMIC-CXR. The fits converge, but in about a third of them an interaction indicator is perfectly separated, so the likelihood is flat along that direction and the coefficient is set by the penalty, which is where implementations differ.

Of the inference harness, the narrative prompt, the binary system message and question template, the chat-template handling and the decoding settings (greedy, no sampling) are retained. Not retained are the run-time resolution of the binary answer tokens, the image selection and preprocessing provenance, and any reproduction gate beyond the MedGemma narrative pass, which was re-run and matched its stored predictions exactly. The analysis code, the retained inference scripts and the per-cell prediction tables will be released, with a persistent identifier added at the proof stage.

\section{Results}
\label{sec:results}

\subsection{Cross Domain Hallucination Heterogeneity}
\label{sec:discovery}

\begin{figure*}[t]
\centering
\includegraphics[width=\textwidth]{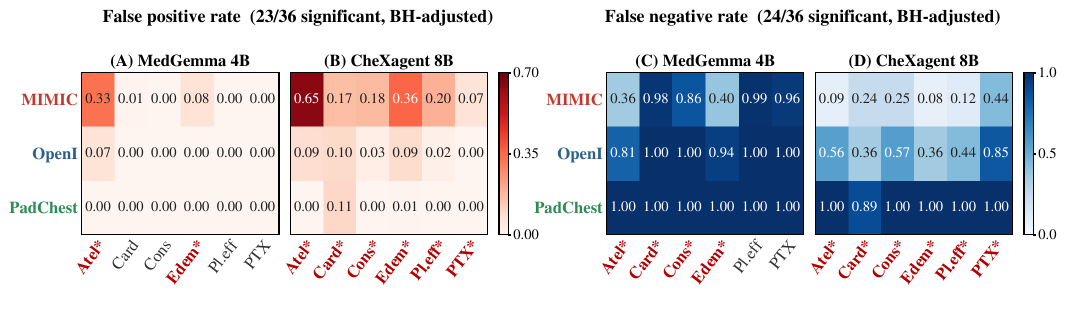}
\caption{Structured cross-institutional heterogeneity in error behavior. (A), (B) false positive rate of MedGemma 4B and CheXagent 8B at each institution; (C), (D) the corresponding false negative rate. Findings are abbreviated (Pl.eff, pleural effusion; PTX, pneumothorax); an asterisk marks a finding with at least one significant pairwise institutional contrast in that panel after Benjamini--Hochberg adjustment over the 36-contrast family for the corresponding metric. All rates are for narrative elicitation. The two systems---model, prompt and parser together---differed despite being evaluated on the same images and reference labels.}
\label{fig:heatmap}
\end{figure*}

Fig.~\ref{fig:heatmap} displays the false positive and false negative rates of the two VLMs across the three deployment domains, and variation was pronounced. For CheXagent, the false positive rate on MIMIC-CXR exceeded that on OpenI by 0.556 for atelectasis ($95\%$ CI $0.395$--$0.714$), by 0.262 for edema and by 0.175 for pleural effusion, with elevated MIMIC-CXR false positive rates for all six findings. For MedGemma, heterogeneity was concentrated in atelectasis ($\Delta=0.261$, $0.042$--$0.487$) and edema ($\Delta=0.075$, $0.046$--$0.107$); the first interval's width reflects a denominator of eighteen negative studies, so effect sizes on rare classes should be read with care.

False negative rates are the more consequential half of the picture and the more extreme. At PadChest both models disagreed with essentially every positive reference label: MedGemma asserted nothing as present in any of its $21{,}862$ evaluable judgments there, and CheXagent reached a false negative rate of 1.00 on five of six findings, cardiomegaly excepted at 0.89. The positive-case denominators run from 6 to 222, so the cells differ greatly in precision. On MIMIC-CXR, at an order of magnitude higher prevalence, CheXagent's ranged from 0.08 to 0.44: the same model tracks the reference standard closely at one institution and not at all at another.

After Benjamini--Hochberg adjustment over all 36 contrasts per metric, 23 false-positive-rate and 24 false-negative-rate contrasts remained significant, and finding-stratified Cochran--Mantel--Haenszel comparisons showed the same direction in all six---descriptively, since several findings can come from one study. That stratification controls which finding is scored, not the case mix within it, so a difference in disease spectrum between an intensive-care and an outpatient population remains a plausible contributor (Section~\ref{sec:limitations}). A study-clustered bootstrap with $10{,}000$ resamples gave a well-defined interval for all 72 cell-level rate differences, of which 54 excluded zero.

\subsection{Distinguishing Model Behavior from Label-Regime Variation}
\label{sec:label_regime}

The three datasets differ not only in patient population but in how their reference labels were produced, so a measured shift could reflect the label regime rather than the model. The confound cannot be removed by design---no public multi-institutional corpus shares one labelling pipeline---but several models were evaluated on identical images with identical labels, which bounds it.

That is not what we observe. CheXagent showed significant false positive rate elevation on MIMIC-CXR for nearly all six findings, MedGemma only on atelectasis and edema; for false-positive rate the models shared 6 of the 17 contrasts significant in at least one model (Jaccard $=6/17=0.353$), so two models scored against the same reference standard disagree on roughly two-thirds of where the degradation falls. Narrative parsing succeeded at different rates, so we repeated the analysis on the $45{,}301$ study-by-finding cells scored for both: the false-positive rejection set and the overlap are unchanged (23 of 36; Jaccard $6/17$). LLaVA-Med contributed no false-positive contrast at all, having asserted no finding as present in any of its $60{,}070$ evaluable judgments (Section~\ref{sec:elicitation}), so its accuracy at each institution equals one minus that institution's reference prevalence.

Because two of the three models fall silent somewhere, we asked whether the images carry recoverable signal at all. On the same images and labels, confidence formed from the affirmative and negative response logits discriminates the reference label at PadChest for CheXagent with an area under the ROC curve of 0.692, above chance for each of the six findings, against 0.468---chance---for LLaVA-Med. The PadChest images therefore carry signal that at least one model recovers, though where any other model's silence originates is not established: no comparable measurement is available for MedGemma there, and the retained evidence does not separate model from prompt, decoding or parser.

Patterns therefore differed across systems on shared image--label pairs, supporting a system-by-environment interaction, but the design cannot remove or quantify contributions from case mix, acquisition or the reference-label pipeline. This also fixes the criterion used throughout---a system is degenerate on a deployment when its rate of asserting a finding does not vary with the image. LLaVA-Med meets it everywhere under both protocols, hence its separate reporting; MedGemma's silence at PadChest is confined to narrative elicitation.

\subsection{Comparing Reliability Estimators}
\label{sec:calibration_main}

Calibration error falls with the target-label budget in every deployment, but the estimators separate in only three of the six: on the other three---MedGemma at OpenI and PadChest, CheXagent at PadChest---every estimator but the constant falls below 0.03 by 200 labels and they differ by less than 0.014, so estimating helps but the choice matters little. Within the structured family the interaction term does what it was designed to do: HD-Cal beats the additive ablation in 22 of 24 settings on both calibration error (median relative reduction $9.9\%$) and Brier score, surviving equal-mass binning and the debiased $\ell_2$ estimator of Kumar \etal~\cite{kumar2019} in 18 and 21 of 24. That advantage is specific to the narrative protocol: on the binary predictions, within one software environment and a four-estimator comparison, it moves from 18 to 5 of 24 settings on Brier. Beta--Binomial empirical Bayes attains lower calibration error than HD-Cal in 23 of the 24 settings. The six deployments are not independent units, two models sharing each institution, so the Brier difference was resampled over the three institutions: the mean gap of $0.0305$ carries a $95\%$ interval of $-0.0005$ to $0.0841$, concentrated at a single site---MIMIC-CXR $+0.0841$, OpenI $+0.0080$, PadChest $-0.0005$---so removing MIMIC-CXR leaves $0.0038$. The advantage of estimating each cell from local counts is thus large in aggregate---mean calibration error 0.054 against 0.114---and heterogeneous across institutions, exploratory rather than confirmatory with three of them. Table~\ref{tab:hdcal_vs_baselines} ranks all ten.

\begin{table}[t]
\centering
\caption{Ten reliability estimators across the 24 external-validation settings, by family. ECE and Brier are averaged over settings; rank is the mean over the ten (1 best). The last column counts settings in which the estimator attains lower ECE than Beta--Binomial.}
\label{tab:hdcal_vs_baselines}
\footnotesize
\setlength{\tabcolsep}{6pt}
\renewcommand{\arraystretch}{1.1}
\begin{tabular}{@{}l c c c r@{}}
\toprule
Estimator & ECE & Brier & Rank & Wins/24 \\
\midrule
\multicolumn{5}{@{}l}{\textit{Constant baseline}}\\
Pooled-rate constant       & 0.109 & 0.118 & 7.29 & 5  \\
\addlinespace[1.5pt]
\multicolumn{5}{@{}l}{\textit{Borrowing structure across institutions}}\\
Additive ($r{+}d{+}k$)     & 0.125 & 0.120 & 8.54 & 2  \\
HD-Cal ($+\,d{\times}k$)   & 0.114 & 0.116 & 7.29 & 1  \\
Weighted HD-Cal            & 0.059 & 0.087 & 5.42 & 4  \\
Hierarchical GLMM          & 0.092 & 0.109 & 4.83 & 8  \\
Adaptive shrinkage (HAS)   & 0.092 & 0.108 & 5.54 & 1  \\
\quad with fixed $\tau$    & 0.093 & 0.109 & 6.29 & 0  \\
\addlinespace[1.5pt]
\multicolumn{5}{@{}l}{\textit{Estimating each cell from target counts}}\\
HAS without source offset  & 0.041 & 0.090 & 2.21 & 15 \\
Target-only logistic       & 0.058 & 0.086 & 4.79 & 7  \\
Beta--Binomial EB          & 0.054 & 0.085 & 2.79 & --- \\
\bottomrule
\end{tabular}
\end{table}

The gain from the interaction term shows the heterogeneity is organized at the institution-by-finding level rather than the marginal one. That conclusion about the \emph{structure} survives; the inference that modelling it is the best way to estimate reliability does not. The separation is also metric-dependent: on calibration error every member of the family borrowing structure across institutions has a higher mean than every member of the family estimating each cell from target counts, whereas on the Brier score the two overlap.

Two qualifications keep this honest. The advantage is aggregate: several estimators attain lower calibration error than Beta--Binomial in a minority of settings, so none of the ten is universally optimal. And the variant leading on calibration error, adaptive shrinkage without its cross-institutional offset, is worse on Brier, which settles it; that estimator was our own attempt to rescue the structured family.

\subsection{Selection Under Strict Institution Hold-Out}
\label{sec:protocol_eval}

Table~\ref{tab:hdcal_vs_baselines} scores estimators one at a time. Algorithm~\ref{alg:protocol} must commit to one before seeing the deployment it will serve, so we ran its selection step under the hold-out rule of Section~\ref{sec:protocol}: 240 model-specific decisions over 240 evaluations, the receiving institution absent from every development fit.

Under fully outer-institution-excluded evaluation, the post-hoc seven-action protocol-compatible selector achieved a mean Brier score of 0.1083, compared with 0.0853 for always using Beta--Binomial and 0.0855 for always using target-only logistic. Against the Set-B per-cell oracle at $0.0799$ that is a relative excess of $35.5\%$; the selector's Brier is $8.0\%$ lower than the pooled-rate constant at $0.1177$, but that compares the complete procedure with a one-rate, no-cell-structure baseline and does not isolate the value of adaptive selection. In 4 of the 240 decisions no non-constant candidate achieved a strictly lower development Brier than the pooled-rate constant, so the gate returned that baseline. The selected action varied across the 240 decisions: Beta--Binomial 79 times, the GLMM 60, target-only logistic 38, HD-Cal 24, weighted HD-Cal 23, additive 12 and the pooled-rate constant 4.

Table~\ref{tab:policies} sets the procedure against fixed policies on identical folds, budgets, repetitions and test sets. It did not outperform the numerically leading eligible fixed policies: Beta--Binomial and the target-only logistic model both remain below it, and we report this step of Algorithm~\ref{alg:protocol} as evaluated and unsupported. The two leaders differ by $0.0003$, less than this family's sensitivity to a change of solver version (Section~\ref{sec:reproducibility}), and each leads in about half the settings, so the primary analysis does not order them.

The matched five-candidate set was fixed after the ten-candidate result was known, the hierarchical and adaptive-shrinkage estimators being unavailable where the binary protocol can be scored, and it carries no selection result: only fixed policies are compared across interfaces (Table~\ref{tab:policies}B), recomputed within a single software environment and so not comparable in absolute terms with panel A. Within it the two leaders change places; HD-Cal and the additive model change from higher Brier than the pooled-rate constant under narrative elicitation to lower Brier under binary, whereas weighted HD-Cal remains lower than the constant under both protocols.

The Beta--Binomial interval was evaluated rather than only constructed, against the estimand a held-out fold can test. At a nominal $95\%$ level the plug-in empirical-Bayes posterior-predictive count interval covered the held-out count in $87.0\%$ of $1{,}876$ cell-evaluations at a mean width of $0.299$, and coverage was lowest where it matters most: $78.2\%$ at MIMIC-CXR against $96.6\%$ at PadChest, and $82.0\%$ for judgments asserting a finding present against $88.8\%$ for absent. The shortfall is not an artefact of the eligibility rule---admitting every evaluable cell gives $87.3\%$---nor of the budget. A discrete equal-tailed interval is conservative by construction, so this reading is the optimistic one. These evaluations overlap heavily, so they are point estimates without an accompanying interval.

\begin{table}[t]
\centering
\caption{The selection step of Algorithm~\ref{alg:protocol} against fixed policies, on identical folds, budgets, repetitions and test sets. (A) narrative elicitation under the strict institution hold-out of Section~\ref{sec:protocol}: 240 model-specific decisions over 240 evaluations, the selector ranging over the seven protocol-compatible actions, with excess measured against the per-cell oracle over those seven---an unattainable lower bound, not an available policy. (B) fixed policies on the post-hoc matched five-candidate set: not pre-registered, recomputed within a single software environment, and so not comparable with (A). The two leading policies change places between the interfaces; the best in each column is bold.}
\label{tab:policies}
\footnotesize
\setlength{\tabcolsep}{4pt}
\renewcommand{\arraystretch}{1.1}
\begin{tabular}{@{}l c c c@{}}
\multicolumn{4}{@{}l}{\textit{(A) Seven protocol-compatible actions, narrative}}\\
\toprule
Policy & Brier & vs pooled & Excess \\
\midrule
Algorithm~\ref{alg:protocol} selection & 0.1083 & $8.0\%$  & $35.5\%$ \\
Always Beta--Binomial                  & 0.0853 & $27.6\%$ & $6.7\%$ \\
Always target-only logistic            & 0.0855 & $27.3\%$ & $7.0\%$ \\
Always weighted HD-Cal                 & 0.0870 & $26.1\%$ & $8.8\%$ \\
Pooled-rate constant                   & 0.1177 & ---      & $47.3\%$ \\
Per-cell oracle (lower bound)          & 0.0799 & $32.1\%$ & --- \\
\bottomrule
\end{tabular}

\vspace{3.5pt}

\begin{tabular}{@{}l c c@{}}
\multicolumn{3}{@{}l}{\textit{(B) Post-hoc matched five-candidate fixed policies}}\\
\toprule
Policy & Narrative & Binary \\
\midrule
Always Beta--Binomial                  & 0.0853 & \textbf{0.0904} \\
Always target-only logistic            & \textbf{0.0827} & 0.0926 \\
Always weighted HD-Cal                 & 0.0870 & 0.0922 \\
Always HD-Cal                          & 0.1243 & 0.1176 \\
Always additive                        & 0.1292 & 0.1106 \\
Pooled-rate constant                   & 0.1177 & 0.1389 \\
\bottomrule
\end{tabular}
\end{table}

\subsection{Hard Domain Calibration and the Difficulty Gap Relationship}
\label{sec:harddomain}

How the structured estimators behave depends on how far the receiving institution departs from the source. For the hardest deployment---MedGemma on MIMIC-CXR, a 37-point source--target accuracy gap, 200 labels---calibration error was 0.366 for the pooled-rate constant, 0.209 for HD-Cal and 0.078 for Beta--Binomial (Table~\ref{tab:hdcal_vs_baselines}, Fig.~\ref{fig:overview}).

Where the gap is small they do not merely fail to help---they harm calibration. For CheXagent on MIMIC-CXR the gap was about 11 points and the pooled-rate constant already well calibrated at 0.113, yet HD-Cal was worse at every budget---0.224 even with 200 labels, and $2.29\times$ worse than the constant when averaged over budgets. Aggregated over the evaluation, HD-Cal is worse than the constant on 9 of 24 settings by calibration error and 7 of 24 by Brier, and the additive baseline is worse at every budget in both CheXagent settings. This is a property of the family: a fixed regularization strength cannot adapt to the source--target gap, under-correcting where it is large and over-correcting where small.

\subsection{Reliability Also Depends on How the Model Is Asked}
\label{sec:elicitation}

\begin{figure}[!htb]
\centering
\includegraphics[width=\columnwidth]{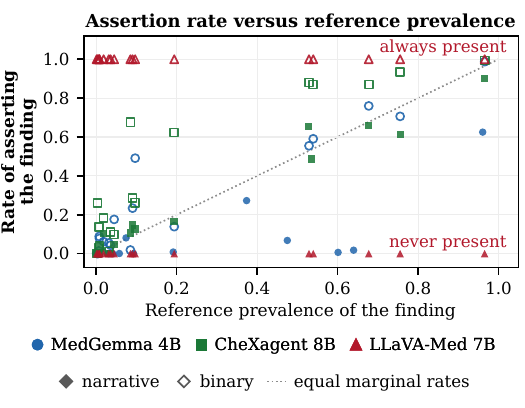}
\caption{Marginal rate at which each system asserts a finding, against that finding's reference prevalence, per institution-by-finding cell. Marker shape denotes the model and fill the elicitation protocol: filled narrative, open binary. The dotted line marks equal marginal rates, not agreement on individual images. LLaVA-Med lies on the horizontal lines at zero and one, which marginal rates alone cannot attribute to the model rather than the prompt or the parser. Narrative elicitation requires four of six findings to parse, so a cell's prevalence can differ marginally between protocols.}
\label{fig:elicitation}
\end{figure}

The analyses above treat a model's judgment as fixed for a given image and finding. It is not. Under binary elicitation the same models on the same images returned substantially different judgments (Fig.~\ref{fig:elicitation}): raw agreement was 89.7\% and 85.1\%, but both protocols are dominated by absent judgments, so that overstates concordance---Cohen's $\kappa$ is 0.200 and 0.518, and accuracy falls by 3.0 and 11.4 points.

For LLaVA-Med the protocols were not merely discordant but opposite: narrative elicitation asserted no finding as present in any of $60{,}070$ evaluable judgments, binary elicitation asserted every one. Its rate of asserting a finding is therefore uninformative about the image, visible in Fig.~\ref{fig:elicitation} as two horizontal lines against reference prevalences spanning almost the full unit interval. Whether this is an intrinsic defect or a matter of prompt fit we cannot say, one prompt having been used throughout.

\subsection{Robustness Checks}
\label{sec:sensitivity}

Two checks bound the conclusions. Uncertain labels were handled three ways---mapped to negative, to positive, or excluded---and the within-family comparison held in $22/24$, $21/24$ and $22/24$ settings; only the false negative count was sensitive to this.

\label{sec:dedup_sensitivity}
MIMIC-CXR was sampled at the study rather than patient level (3{,}066 evaluable studies from 2{,}390 patients), though repeated patients cannot leak across institutions. Restricting to one study per patient under two deduplication rules preserved the within-family advantage on Brier in $22/24$ settings and on calibration error in $19/24$ and $20/24$; a size-matched control retaining repeats yielded $20$ to $23$ wins across five draws. The movement tracks sample size rather than patient repetition, so win counts of this kind carry a jitter of one to two settings.

Four checks within a single software environment (Section~\ref{sec:reproducibility})---leave-one-finding-out, $75\%$ study subsampling, study-disjoint splitting, and an oracle over five HD-Cal penalties spanning four orders of magnitude---did not explain the family difference; the penalty used throughout was not the one most favourable to the estimators working from target counts.

\subsection{Single-Reader Audit of High-Risk Cells}
\label{sec:audit_results}

Of 900 reader annotations, 774 study--finding pairs had at least one evaluable model record under Section~III-B; excluding uncertain reader labels and collapsing model-expanded rows left 707 unique pairs. Their moderate agreement with dataset labels (Cohen's $\kappa=0.502$) implicates label provenance in some discrepancies. Among 1,305 model-expanded records, top- and bottom-risk cells had reader-flagged error rates of $21.3\%$ and $13.3\%$, respectively, and came from 127 and 132 studies; cluster-aware procedures disagreed. This error-enriched, single-reader audit of 150 studies is descriptive, not a powered reader study.

\section{Discussion}
\label{sec:discussion}

\subsection{What the Strict Design Changes}
\label{sec:significance}

Excluding the receiving institution from development altogether is not exotic: it is the design a reader would assume was in force, and the position a receiving site is actually in. Imposing it turns the selection result negative---a caution about how deployment-time evaluations are reported rather than about this estimator set. Strict institution exclusion turns selection negative because each decision is learned from only two same-model pseudo-target folds, each fitted with one rather than the deployment's two source institutions; ranking near-equivalent estimators under that mismatch can cost more than a fixed policy.

\subsection{What Transfers, and What Does Not}
\label{sec:simplicity}

The two policies that led are both classical. With agreement indexed by institution, finding and prediction direction, the problem is a few dozen binomial rates with no instance-level covariate, so estimating each cell from its own counts is close to what that information permits---which is why our own adaptive-shrinkage variant failed to beat it. That measuring at the receiving site earns its cost shows in one contrast: agreement carried over from the development sites would credit a model with $97.0\%$ where it was $60.2\%$ at the receiving site.

What the evidence does not support is a recommendation between them: the two are separated by less than the divergence the same logistic family shows between solver versions (Section~\ref{sec:reproducibility}), and change places again in the matched-set sensitivity analysis. Their advantage over structured pooling rests on one of the three institutions, and the interval on the difference includes zero once the clusters are institutions rather than deployments (Section~\ref{sec:calibration_main}); three clusters could not establish otherwise. Nor is the ordering invariant to the interface: on the matched five-candidate set HD-Cal and the additive model move from higher to lower Brier than the constant predictor between the two interfaces, while weighted HD-Cal stays lower than it under both---a sensitivity analysis, not a second confirmation.

The posterior-predictive interval undercovered even after estimand alignment, especially in the hardest settings; receiving sites should therefore re-estimate agreement for each interface, benchmark against simple fixed policies and the constant predictor, and treat these intervals as approximate.

\subsection{What Instance-Level Signal Would Add}
\label{sec:logit_outlook}

The results suggest a limitation of the evaluated categorical cell representation and estimator set; instance-level confidence may provide additional discrimination. Every estimator here is bounded by categorical inputs to one value per cell, so none can rank judgments within a cell. Token log-probabilities lift that bound: on the binary pass, confidence from the affirmative and negative logits predicts whether a judgment is correct with an area under the ROC curve of 0.796 for MedGemma and 0.743 for CheXagent, reaching 0.898 for pneumothorax---an ordering no cell-level estimator can produce, though at chance for the degenerate model.

\section{Limitations}
\label{sec:limitations}

Several limitations bound the interpretation. The evidence comes from three public retrospective corpora rather than a live service, so institution-level inference rests on three clusters, too few for any interval here to be confirmatory. Those corpora use different reference-label pipelines, and what is estimated is agreement with those labels on the subset carrying a definite label; reader--dataset agreement was $\kappa=0.502$, so $97.0\%$ or $60.2\%$ is agreement with a label of moderate quality, not with clinical truth. Section~\ref{sec:label_regime} bounds that entanglement without removing it. The heterogeneity reported here is accordingly a joint property of the evaluated system and its corpus and reference-label environment; the design cannot separate institution, case mix, acquisition and labelling pipeline. Fig.~\ref{fig:heatmap} is also reference-conditioned---error rates given the reference label---a diagnostic-error view rather than a direct analysis of the prediction-conditioned estimand of (\ref{eq:hdcal}).

Four constraints are specific to the analyses. Fit and test rows never share a study under the disjoint split, but no patient identifier is available, so patient-level disjointness is untested. The selector's seven-action set is post-hoc: the adaptive-shrinkage variants are excluded because their implementation is undefined when a development fold supplies one source institution instead of two---structural, not a verdict on accuracy---and they stay in the fixed-policy benchmark. That asymmetry makes development and deployment different tasks, so a development ranking need not carry over; with three institutions admitting one such nesting, nothing here shows adaptive selection fails in general. The narrative--binary contrast runs on fixed policies from a set fixed after the ten-candidate result was known and recomputed in a different software environment: exploratory and hypothesis-generating, not independent confirmation. The intervals stay below their nominal level under the estimand a held-out fold can test, and at non-zero budgets the pooled-rate constant is itself updated with the target shot, so a margin over it compares the complete procedure with a one-rate, no-cell-structure baseline rather than isolating either local labelling or adaptive selection.

Two further gaps are documentary and clinical. The narrative prompt, the binary template and the decoding settings are retained; the run-time resolution of the binary answer tokens, the image selection and preprocessing provenance, and any reproduction gate beyond the MedGemma narrative pass are not, so the binary results are exploratory throughout and the third VLM's uniform outputs are reported as parsed behaviour of the evaluated prompt--template--parser configuration rather than attributed to the model. The audit is one reader over 150 studies under error-enriched sampling: descriptive, and supporting no claim of clinical benefit.

\section{Conclusion}
\label{sec:conclusion}

What can be concluded about estimating reference agreement at a receiving institution depends on how the estimation is validated. Under evaluation excluding the receiving institution from development entirely, adaptively choosing among the seven estimators that design admits did not improve on simple fixed alternatives. The two that led were too close, and too unstable across settings and interfaces, to support a universal recommendation; their advantage over structured pooling rested on one institution, and an estimand-aligned posterior-predictive interval did not attain its nominal level. Receiving institutions should re-evaluate reference agreement for their own site and interface; nothing established here concerns clinical correctness or safe deployment.

\def\refname{References}
\bibliographystyle{IEEEtran}
\bibliography{hdcal_references}

@article{medgemma,
  author        = {Sellergren, Andrew and others},
  title         = {{MedGemma Technical Report}},
  journal       = {arXiv preprint arXiv:2507.05201},
  year          = {2025},
  eprint        = {2507.05201},
  archivePrefix = {arXiv},
  primaryClass  = {cs.AI},
  doi           = {10.48550/arXiv.2507.05201}
}

@article{chexagent,
  author        = {Chen, Zhihong and others},
  title         = {{A Vision-Language Foundation Model to Enhance Efficiency of Chest X-ray Interpretation}},
  journal       = {arXiv preprint arXiv:2401.12208},
  year          = {2024},
  eprint        = {2401.12208},
  archivePrefix = {arXiv},
  doi           = {10.48550/arXiv.2401.12208}
}

@article{unichest,
  author  = {Dai, T. and Zhang, R. and Hong, F. and Yao, J. and Zhang, Y. and Wang, Y.},
  title   = {{UniChest: Conquer-and-Divide Pre-training for Multi-Source Chest X-Ray Classification}},
  journal = {IEEE Trans. Med. Imag.},
  volume  = {43},
  number  = {8},
  pages   = {2901--2912},
  year    = {2024},
  doi     = {10.1109/TMI.2024.3381123}
}

@inproceedings{llava_med,
  author    = {Li, C. and others},
  title     = {{LLaVA-Med: Training a Large Language-and-Vision Assistant for Biomedicine in One Day}},
  booktitle = {Proc. Adv. Neural Inf. Process. Syst. (Datasets and Benchmarks Track)},
  volume    = {36},
  year      = {2023}
}

@article{multimodal_med_foundation,
  author  = {Moor, M. and others},
  title   = {{Foundation models for generalist medical artificial intelligence}},
  journal = {Nature},
  volume  = {616},
  number  = {7956},
  pages   = {259--265},
  year    = {2023},
  doi     = {10.1038/s41586-023-05881-4}
}

@article{radfm,
  author  = {Wu, C. and Zhang, X. and Zhang, Y. and Hui, H. and Wang, Y. and Xie, W.},
  title   = {{Towards generalist foundation model for radiology by leveraging web-scale 2D{\&}3D medical data}},
  journal = {Nat. Commun.},
  volume  = {16},
  number  = {1},
  pages   = {7866},
  year    = {2025},
  doi     = {10.1038/s41467-025-62385-7}
}

@inproceedings{rextrust,
  author    = {Hardy, R. and Kim, S. E. and Ro, D. H. and Rajpurkar, P.},
  title     = {{ReXTrust: A Model for Fine-Grained Hallucination Detection in AI-Generated Radiology Reports}},
  booktitle = {Proc. AAAI Bridge Program AI Med. Healthcare},
  series    = {Proc. Mach. Learn. Res.},
  volume    = {281},
  pages     = {173--182},
  year      = {2025}
}

@inproceedings{radflag,
  author    = {Zhang, S. and Sambara, S. and Banerjee, O. and Acosta, J. N. and Fahrner, L. J. and Rajpurkar, P.},
  title     = {{RadFlag: A Black-Box Hallucination Detection Method for Medical Vision Language Models}},
  booktitle = {Proc. Mach. Learn. Health Symp.},
  series    = {Proc. Mach. Learn. Res.},
  volume    = {259},
  pages     = {1087--1103},
  year      = {2025}
}

@article{medvh,
  author  = {Gu, Z. and Chen, J. and Liu, F. and Yin, C. and Zhang, P.},
  title   = {{MedVH: Toward Systematic Evaluation of Hallucination for Large Vision Language Models in the Medical Context}},
  journal = {Adv. Intell. Syst.},
  volume  = {8},
  number  = {1},
  pages   = {2500255},
  year    = {2026},
  doi     = {10.1002/aisy.202500255}
}

@article{crest,
  author  = {Guan, H. and others},
  title   = {{A Clinically-Informed Framework for Evaluating Vision-Language Models in Radiology Report Generation: Taxonomy of Errors and Risk-Aware Metric}},
  journal = {AMIA Annu. Symp. Proc.},
  volume  = {2024},
  pages   = {383--392},
  year    = {2025}
}

@inproceedings{cares,
  author    = {Xia, P. and others},
  title     = {{CARES: a comprehensive benchmark of trustworthiness in medical vision language models}},
  booktitle = {Proc. Adv. Neural Inf. Process. Syst. (Datasets and Benchmarks Track)},
  volume    = {37},
  year      = {2024}
}

@article{domainshift_medical,
  author  = {Zech, J. R. and Badgeley, M. A. and Liu, M. and Costa, A. B. and Titano, J. J. and Oermann, E. K.},
  title   = {{Variable generalization performance of a deep learning model to detect pneumonia in chest radiographs: a cross-sectional study}},
  journal = {PLoS Med.},
  volume  = {15},
  number  = {11},
  pages   = {e1002683},
  year    = {2018},
  doi     = {10.1371/journal.pmed.1002683}
}

@article{finlayson2021,
  author  = {Finlayson, S. G. and others},
  title   = {{The clinician and dataset shift in artificial intelligence}},
  journal = {New England J. Med.},
  volume  = {385},
  number  = {3},
  pages   = {283--286},
  year    = {2021},
  doi     = {10.1056/NEJMc2104626}
}

@inproceedings{guo2017calibration,
  author    = {Guo, C. and Pleiss, G. and Sun, Y. and Weinberger, K. Q.},
  title     = {{On calibration of modern neural networks}},
  booktitle = {Proc. Int. Conf. Mach. Learn.},
  series    = {Proc. Mach. Learn. Res.},
  volume    = {70},
  pages     = {1321--1330},
  year      = {2017}
}

@inproceedings{naeini2015,
  author    = {Naeini, M. P. and Cooper, G. F. and Hauskrecht, M.},
  title     = {{Obtaining Well Calibrated Probabilities Using Bayesian Binning}},
  booktitle = {Proc. AAAI Conf. Artif. Intell.},
  volume    = {29},
  number    = {1},
  pages     = {2901--2907},
  year      = {2015},
  doi       = {10.1609/aaai.v29i1.9602}
}

@inproceedings{kumar2019,
  author    = {Kumar, A. and Liang, P. and Ma, T.},
  title     = {{Verified uncertainty calibration}},
  booktitle = {Proc. Adv. Neural Inf. Process. Syst.},
  volume    = {32},
  pages     = {3792--3803},
  year      = {2019}
}

@inproceedings{ovadia2019,
  author    = {Ovadia, Y. and others},
  title     = {{Can you trust your model's uncertainty? Evaluating predictive uncertainty under dataset shift}},
  booktitle = {Proc. Adv. Neural Inf. Process. Syst.},
  volume    = {32},
  pages     = {13969--13980},
  year      = {2019}
}

@inproceedings{geifman_selective,
  author    = {Geifman, Y. and El-Yaniv, R.},
  title     = {{Selective classification for deep neural networks}},
  booktitle = {Proc. Adv. Neural Inf. Process. Syst.},
  volume    = {30},
  pages     = {4885--4894},
  year      = {2017}
}

@inproceedings{selectivenet,
  author    = {Geifman, Y. and El-Yaniv, R.},
  title     = {{SelectiveNet: a deep neural network with an integrated reject option}},
  booktitle = {Proc. Int. Conf. Mach. Learn.},
  series    = {Proc. Mach. Learn. Res.},
  volume    = {97},
  pages     = {2151--2159},
  year      = {2019}
}

@article{conformal,
  author  = {Angelopoulos, A. N. and Bates, S.},
  title   = {{Conformal Prediction: A Gentle Introduction}},
  journal = {Found. Trends Mach. Learn.},
  volume  = {16},
  number  = {4},
  pages   = {494--591},
  year    = {2023},
  doi     = {10.1561/2200000101}
}

@inproceedings{conformal_risk,
  author    = {Angelopoulos, A. N. and Bates, S. and Fisch, A. and Lei, L. and Schuster, T.},
  title     = {{Conformal risk control}},
  booktitle = {Proc. Int. Conf. Learn. Represent.},
  year      = {2024}
}

@article{mimiccxr,
  author  = {Johnson, A. E. W. and others},
  title   = {{MIMIC-CXR, a de-identified publicly available database of chest radiographs with free-text reports}},
  journal = {Sci. Data},
  volume  = {6},
  number  = {1},
  pages   = {317},
  year    = {2019},
  doi     = {10.1038/s41597-019-0322-0}
}

@article{openi,
  author  = {Demner-Fushman, D. and others},
  title   = {{Preparing a collection of radiology examinations for distribution and retrieval}},
  journal = {J. Amer. Med. Inform. Assoc.},
  volume  = {23},
  number  = {2},
  pages   = {304--310},
  year    = {2016},
  doi     = {10.1093/jamia/ocv080}
}

@article{padchest,
  author  = {Bustos, A. and Pertusa, A. and Salinas, J.-M. and de la Iglesia-Vay{\'a}, M.},
  title   = {{PadChest: A large chest x-ray image dataset with multi-label annotated reports}},
  journal = {Med. Image Anal.},
  volume  = {66},
  pages   = {101797},
  year    = {2020},
  doi     = {10.1016/j.media.2020.101797}
}

@inproceedings{chexpert_labeler,
  author    = {Irvin, J. and others},
  title     = {{CheXpert: a large chest radiograph dataset with uncertainty labels and expert comparison}},
  booktitle = {Proc. AAAI Conf. Artif. Intell.},
  volume    = {33},
  number    = {01},
  pages     = {590--597},
  year      = {2019},
  doi       = {10.1609/aaai.v33i01.3301590}
}

@article{claim2024,
  author  = {Tejani, A. S. and others},
  title   = {{Checklist for Artificial Intelligence in Medical Imaging (CLAIM): 2024 Update}},
  journal = {Radiol. Artif. Intell.},
  volume  = {6},
  number  = {4},
  pages   = {e240300},
  year    = {2024},
  doi     = {10.1148/ryai.240300}
}

@article{tripod_ai,
  author  = {Collins, G. S. and others},
  title   = {{TRIPOD+AI statement: updated guidance for reporting clinical prediction models that use regression or machine learning methods}},
  journal = {BMJ},
  volume  = {385},
  pages   = {e078378},
  year    = {2024},
  doi     = {10.1136/bmj-2023-078378}
}

@article{sadanadan2026,
  author        = {Sadanandan, Binesh and Behzadan, Vahid},
  title         = {{Predictive Entropy Links Calibration and Paraphrase Sensitivity in Medical Vision-Language Models}},
  journal       = {arXiv preprint arXiv:2604.08941},
  year          = {2026},
  eprint        = {2604.08941},
  archivePrefix = {arXiv},
  doi           = {10.48550/arXiv.2604.08941}
}

@article{bh1995,
  author  = {Benjamini, Y. and Hochberg, Y.},
  title   = {{Controlling the false discovery rate: a practical and powerful approach to multiple testing}},
  journal = {J. Roy. Statist. Soc. B},
  volume  = {57},
  number  = {1},
  pages   = {289--300},
  year    = {1995},
  doi     = {10.1111/j.2517-6161.1995.tb02031.x}
}

@article{sklearn,
  author  = {Pedregosa, F. and others},
  title   = {{Scikit-learn: machine learning in Python}},
  journal = {J. Mach. Learn. Res.},
  volume  = {12},
  pages   = {2825--2830},
  year    = {2011}
}

\end{document}